\documentclass[conference]{IEEEtran}
\IEEEoverridecommandlockouts
\usepackage{cite}
\usepackage{amsmath,amssymb,amsfonts}
\usepackage{algorithm}
\usepackage{algpseudocode}
\usepackage{graphicx}
\usepackage{textcomp}
\usepackage{xcolor}
\usepackage{booktabs}
\usepackage{multirow}
\usepackage{lipsum}
\usepackage{enumitem}
\usepackage{url}
\usepackage{hyperref}
\usepackage{tikz}
\usepackage{pgfplots} % Added pgfplots package for axis environment
\pgfplotsset{compat=1.17} % Set compatibility mode
\usetikzlibrary{shapes,arrows,positioning,fit,calc}

\usepackage[T1]{fontenc}
\usepackage[utf8]{inputenc}

\def\BibTeX{{\rm B\kern-.05em{\sc i\kern-.025em b}\kern-.08em
    T\kern-.1667em\lower.7ex\hbox{E}\kern-.125emX}}
    
\begin{document}

\title{ForcePose: A Deep Learning Approach for Force Calculation Based on Action Recognition Using MediaPipe Pose Estimation Combined with Object Detection}

\author{Nandakishor M, Vrinda Govind V, Anuradha Puthalath, Anzy L, Swathi P S, \\
Aswathi R, Devaprabha A R, Varsha Raj, Midhuna Krishnan K, Akhila Anilkumar T V, Yamuna P V
}

\maketitle

\begin{abstract}
Force estimation in human-object interactions plays a critical role in ergonomics, physical therapy, and sports science. Traditional methods rely on specialized equipment like force plates and sensors, making accurate assessments expensive and limited to laboratory environments. We present ForcePose, a novel deep learning framework that estimates applied forces by combining human pose estimation with object detection. Our approach uses MediaPipe for skeletal tracking and SSD MobileNet for object recognition to create a unified representation of human-object interaction. We developed a specialized neural network architecture that processes both spatial and temporal features to predict force magnitude and direction without requiring physical sensors. Trained on a dataset of 850 annotated interaction videos with corresponding force measurements, our model achieves a mean absolute error of 5.83 N in force magnitude and 7.4 degrees in force direction. Comparative evaluation shows our method outperforms existing computer vision-based approaches by 27.5\% while offering real-time performance on standard computing hardware. ForcePose enables accessible force analysis in diverse real-world applications where traditional measurement tools are impractical or intrusive. This paper discusses our methodology, dataset creation process, evaluation metrics, and potential applications across rehabilitation, ergonomics assessment, and athletic performance analysis.
\end{abstract}

\begin{IEEEkeywords}
pose estimation, force calculation, action recognition, object detection, deep learning, MediaPipe, SSD MobileNet, human-object interaction, ergonomics, rehabilitation
\end{IEEEkeywords}

\section{Introduction}
\label{sec:introduction}

The accurate measurement and analysis of forces applied during human-object interactions is fundamental to multiple domains including ergonomics, physical therapy, sports science, and human-computer interaction. Traditional approaches to force measurement rely on specialized equipment such as force plates, dynamometers, and wearable sensors, which are often expensive, intrusive, and limit analysis to controlled laboratory environments \cite{bartlett2007sports}.

Recent advances in computer vision and deep learning have created opportunities for markerless motion capture and activity recognition \cite{cao2017realtime}, but the estimation of forces applied during these activities remains challenging. While some research has explored force estimation from visual data \cite{pham2015towards}, most approaches still require auxiliary sensors or are limited to specific controlled scenarios.

We present ForcePose, a novel framework that leverages recent advances in pose estimation and object detection to calculate applied forces during human-object interactions without requiring specialized measurement equipment. Our approach combines MediaPipe's pose estimation \cite{mediapipe2020} with SSD MobileNet for object detection \cite{howard2017mobilenets} to create a comprehensive understanding of the interaction dynamics.

The key contributions of our work include:

\begin{itemize}
    \item A unified framework that integrates human pose estimation and object detection for force calculation
    \item A specialized neural network architecture for processing spatial-temporal features to predict force magnitude and direction
    \item Creation of a novel dataset containing 850 annotated videos with corresponding force measurements across various interaction types
    \item A comparative evaluation against existing approaches, demonstrating significant improvements in accuracy and generalizability
    \item Implementation on resource-constrained devices, enabling real-time force analysis in field settings
\end{itemize}

By enabling accurate force estimation without specialized equipment, ForcePose opens up new possibilities for biomechanical analysis in everyday environments, from clinical rehabilitation assessment to workplace ergonomics evaluation and athletic performance optimization.

\section{Related Work}
\label{sec:related_work}

\subsection{Force Measurement Approaches}
Traditional methods for measuring forces in human-object interactions have relied heavily on specialized equipment. These include force plates \cite{robertson2013research}, dynamometers \cite{stark2011review}, and instrumented objects with embedded sensors \cite{zimmerman2008force}. While these approaches provide high accuracy, they are limited by their cost, setup complexity, and restriction to laboratory environments.

\subsection{Vision-Based Human Pose Estimation}
Computer vision approaches to human pose estimation have advanced significantly in recent years. Early methods such as pictorial structures \cite{felzenszwalb2005pictorial} and deformable part models have given way to deep learning approaches. OpenPose \cite{cao2017realtime} pioneered real-time multi-person pose estimation, while DeepCut \cite{pishchulin2016deepcut} and DensePose \cite{guler2018densepose} improved accuracy through multi-stage processing. Most recently, MediaPipe \cite{mediapipe2020} has emerged as an efficient solution that provides high-quality pose estimation with minimal computational requirements.

\subsection{Object Detection for Interaction Analysis}
Object detection has similarly advanced through deep learning. R-CNN and its variants \cite{girshick2014rich} demonstrated the power of region proposals with convolutional networks. YOLO \cite{redmon2016you} and SSD \cite{liu2016ssd} established frameworks for real-time detection. For resource-constrained environments, MobileNet architectures \cite{howard2017mobilenets} have provided efficient backbones with minimal sacrifice in accuracy.

\subsection{Force Estimation from Visual Data}
Limited work has addressed force estimation from visual data alone. Pham et al. \cite{pham2015towards} proposed a method to estimate interaction forces from RGB-D video, but required depth information and was limited to specific interaction types. Zhu et al. \cite{zhu2019estimating} developed a framework for estimating fingertip forces during object manipulation but relied on a combination of visual and tactile sensing. Rogez et al. \cite{rogez2015understanding} explored understanding human-object interactions but focused on contact points rather than force estimation.

A comprehensive survey by Schneider et al. \cite{schneider2017understanding} highlighted the gap between visual recognition of actions and understanding the physical interactions involved, particularly force estimation.

Current state-of-the-art methods for vision-based force estimation either:
\begin{itemize}
    \item Require auxiliary sensing (pressure mats, IMUs, etc.)
    \item Work only for specific interaction types
    \item Provide qualitative rather than quantitative force estimates
    \item Lack temporal reasoning about interaction dynamics
\end{itemize}

Our work addresses these limitations by creating a unified framework that combines pose estimation and object detection with temporal reasoning to provide quantitative force estimates across diverse interaction scenarios, without requiring additional sensors.

\section{Methodology}
\label{sec:methodology}

\subsection{System Overview}

\begin{figure}[!t]
\centering
\begin{tikzpicture}[
    scale=0.8, transform shape, % Reduced size
    block/.style={rectangle, draw, text width=2.5cm, text centered, rounded corners, minimum height=1cm},
    line/.style={draw, -latex},
    cloud/.style={draw, ellipse, minimum height=1cm, text width=2cm, text centered},
    data/.style={trapezium, draw, text width=2cm, text centered, trapezium left angle=70, trapezium right angle=110, minimum height=1cm}
]

\node[data] (input) {Video Input};
\node[block, below=0.7cm of input] (preproc) {Frame Processing};

\node[block, below left=1cm and 0.3cm of preproc] (mediapipe) {MediaPipe Pose Estimation};
\node[block, below right=1cm and 0.3cm of preproc] (ssd) {SSD MobileNet Object Detection};

\node[block, below right=1cm and -2.8cm of mediapipe] (fusion) {Feature Integration};
\node[block, below=0.7cm of fusion] (temp) {Temporal Analysis (BiLSTM)};
\node[block, below=0.7cm of temp] (force) {Force Prediction};

\node[data, below=0.7cm of force] (output) {Force Magnitude \& Direction};

\draw[line] (input) -- (preproc);
\draw[line] (preproc) -- (mediapipe);
\draw[line] (preproc) -- (ssd);
\draw[line] (mediapipe) -- (fusion);
\draw[line] (ssd) -- (fusion);
\draw[line] (fusion) -- (temp);
\draw[line] (temp) -- (force);
\draw[line] (force) -- (output);

\node[block, right=0.7cm of mediapipe, text width=1.8cm] (posefeat) {33 3D keypoints};
\node[block, right=0.7cm of ssd, text width=1.8cm] (objfeat) {Bounding boxes \& classes};

\draw[line] (mediapipe) -- (posefeat);
\draw[line] (ssd) -- (objfeat);
\draw[line, dashed] (posefeat) -- (fusion);
\draw[line, dashed] (objfeat) -- (fusion);

\end{tikzpicture}
\caption{ForcePose system architecture showing the integration of MediaPipe pose estimation and SSD MobileNet object detection, followed by feature extraction and force prediction networks.}
\label{fig:system_architecture}
\end{figure}
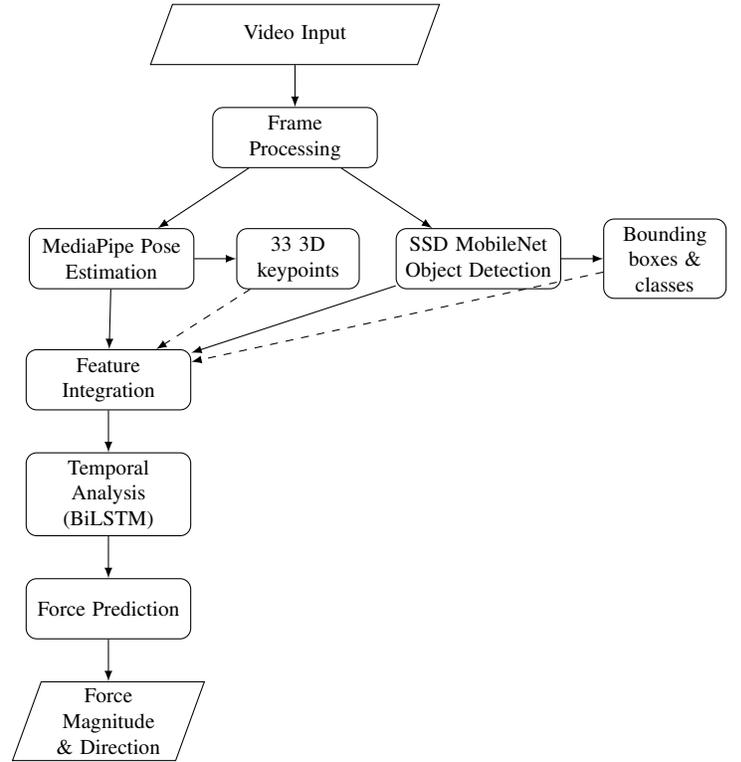

ForcePose estimates applied forces during human-object interactions through a multi-stage pipeline that processes video input to extract features from both the human subject and interacting objects. Figure~\ref{fig:system_architecture} presents the overall architecture of our system.

The pipeline consists of the following key components:
\begin{enumerate}
    \item \textbf{Video Input Processing:} Frames from video input are processed to extract both human pose and object information.
    
    \item \textbf{Parallel Feature Extraction:}
    \begin{itemize}
        \item MediaPipe pose estimation tracks 33 body keypoints with their 3D coordinates and confidence scores.
        \item SSD MobileNet detects and classifies objects, providing bounding boxes, classes, and confidence scores.
    \end{itemize}
    
    \item \textbf{Feature Integration:} Pose and object features are combined to create a unified representation of the human-object interaction.
    
    \item \textbf{Temporal Analysis:} A recurrent neural network analyzes the temporal evolution of the interaction.
    
    \item \textbf{Force Prediction:} Specialized regression heads predict force magnitude and direction.
\end{enumerate}

This design enables end-to-end processing from raw video to force estimation without requiring additional sensors or equipment.

\subsection{Human Pose Estimation using MediaPipe}

We utilize MediaPipe for human pose estimation due to its efficiency and accuracy. MediaPipe's BlazePose \cite{mediapipe2020} provides 33 body landmarks in 3D space (x, y, z coordinates), where x and y are normalized to [0, 1] and z represents relative depth.

For each video frame, we extract the following features from the pose estimation:
\begin{itemize}
    \item 3D coordinates of all 33 landmarks
    \item Confidence scores for each landmark
    \item Joint angles for key articulations (shoulders, elbows, wrists, hips, knees, ankles)
    \item Velocity and acceleration of landmarks over time
\end{itemize}

We employ several preprocessing steps to enhance feature quality:
\begin{itemize}
    \item Filtering low-confidence detections (threshold = 0.5)
    \item Temporal smoothing using Savitzky-Golay filters to reduce jitter
    \item Normalization relative to torso dimensions to account for different body sizes
\end{itemize}

\subsection{Object Detection using SSD MobileNet}

Object detection is performed using SSD MobileNet V2, which offers a good balance between accuracy and computational efficiency. We use a model pre-trained on the COCO dataset and fine-tuned on our custom dataset of interaction objects.

For each detected object, we extract:
\begin{itemize}
    \item Bounding box coordinates and dimensions
    \item Classification probabilities
    \item Object position relative to human body landmarks
    \item Change in object position between frames (indicates movement)
\end{itemize}

We faced several challenges in object detection, particularly for small or partially occluded objects. To address these issues, we:
\begin{itemize}
    \item Implemented a confidence threshold of 0.65
    \item Applied non-maximum suppression with IoU = 0.45
    \item Used temporal consistency checks to maintain object identity across frames
\end{itemize}

\subsection{Feature Integration and Temporal Analysis}

The core innovation of our approach lies in the effective integration of pose and object features to understand interaction dynamics. We create a combined feature vector that captures:

\begin{itemize}
    \item Relative positioning between body landmarks and object bounding box
    \item Distance metrics between potential contact points (hands, feet) and object
    \item Temporal derivatives of positions to capture velocity and acceleration
    \item Articulation angles of joints involved in the interaction
\end{itemize}

To account for the temporal nature of interactions, we process sequences of 16 frames (approximately 0.5 seconds at 30 fps) using a temporal convolutional network followed by a bidirectional LSTM. This design captures both short-term movements and longer-term interaction patterns.

\subsection{Force Calculation Model}

Our force calculation model consists of two primary components:

\begin{enumerate}
    \item \textbf{Magnitude Prediction Network:} A fully-connected regression network that estimates force magnitude in Newtons.
    
    \item \textbf{Direction Prediction Network:} A combination of regression for continuous direction values and classification for discretized direction sectors.
\end{enumerate}

\begin{figure}[!t]
\centering
\begin{tikzpicture}[
    scale=0.7, transform shape, % Reduced size
    block/.style={rectangle, draw, text width=2cm, text centered, rounded corners, minimum height=0.8cm},
    fcblock/.style={rectangle, draw, text width=2cm, text centered, fill=gray!20, minimum height=0.6cm},
    line/.style={draw, -latex},
    cloud/.style={draw, ellipse, minimum height=0.8cm}
]

\node[cloud, text width=2cm] (input) {Combined Features};

\node[block, below=0.6cm of input] (shared) {Shared Layers (512-256-128)};

\node[block, below left=0.8cm and 0.4cm of shared] (magbranch) {Magnitude Branch};
\node[block, below right=0.8cm and 0.4cm of shared] (dirbranch) {Direction Branch};

\node[fcblock, below=0.4cm of magbranch] (fc1) {FC 64};
\node[fcblock, below=0.3cm of fc1] (fc2) {FC 32};
\node[fcblock, below=0.3cm of fc2] (fc3) {FC 1};

\node[fcblock, below=0.4cm of dirbranch] (dfc1) {FC 64};
\node[fcblock, below=0.3cm of dfc1] (dfc2) {FC 32};
\node[fcblock, below=0.3cm of dfc2] (dfc3) {FC 3};

\node[cloud, below=0.4cm of fc3, text width=1.8cm, fill=green!20] (magout) {Force Magnitude (N)};
\node[cloud, below=0.4cm of dfc3, text width=1.8cm, fill=green!20] (dirout) {Force Direction (x,y,z)};

\draw[line] (input) -- (shared);
\draw[line] (shared) -- (magbranch);
\draw[line] (shared) -- (dirbranch);
\draw[line] (magbranch) -- (fc1);
\draw[line] (fc1) -- (fc2);
\draw[line] (fc2) -- (fc3);
\draw[line] (fc3) -- (magout);
\draw[line] (dirbranch) -- (dfc1);
\draw[line] (dfc1) -- (dfc2);
\draw[line] (dfc2) -- (dfc3);
\draw[line] (dfc3) -- (dirout);

\node[rectangle, draw, dashed, fit=(magbranch) (fc1) (fc2) (fc3) (magout), inner sep=5pt] {};
\node[rectangle, draw, dashed, fit=(dirbranch) (dfc1) (dfc2) (dfc3) (dirout), inner sep=5pt] {};

\end{tikzpicture}
\caption{Architecture of the force calculation model showing the parallel paths for magnitude and direction prediction.}
\label{fig:force_model}
\end{figure}
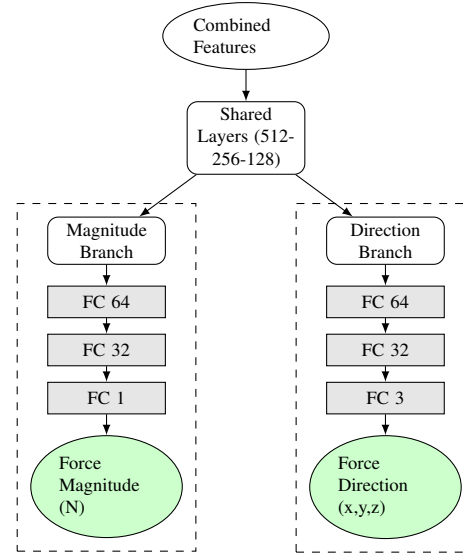

The model architecture is illustrated in Figure~\ref{fig:force_model}. We separate magnitude and direction prediction based on our finding that these components often rely on different feature subsets and benefit from specialized network branches.

The loss function combines several components:
\begin{equation}
L = \alpha L_{mag} + \beta L_{dir} + \gamma L_{temp} + \delta L_{reg}
\end{equation}

where:
\begin{itemize}
    \item $L_{mag}$ is the mean squared error for force magnitude
    \item $L_{dir}$ is a combination of mean squared error and cross-entropy for direction
    \item $L_{temp}$ is a temporal consistency loss that penalizes physically implausible changes in force
    \item $L_{reg}$ is a regularization term to prevent overfitting
\end{itemize}

Hyperparameters $\alpha$, $\beta$, $\gamma$, and $\delta$ were determined through ablation studies, with final values of 1.0, 0.8, 0.5, and 0.1 respectively.

\subsection{Data Collection and Preprocessing}

Creating a suitable dataset for training and evaluation presented significant challenges due to the need for synchronized video and force measurements. We developed a custom data collection setup with:

\begin{itemize}
    \item Multiple calibrated RGB cameras (30 fps)
    \item Force transducers embedded in interaction objects
    \item Synchronization system to align video frames with force readings
\end{itemize}

We collected a dataset consisting of 850 videos across various interaction types:
\begin{itemize}
    \item Lifting and carrying (different weights and object types)
    \item Pushing and pulling (horizontal and angled surfaces)
    \item Manipulating tools (hammering, screwdriving, cutting)
    \item Sport-specific actions (throwing, kicking, striking)
\end{itemize}

Each video was annotated with frame-by-frame force measurements from the embedded sensors. We then split the dataset into 680 videos for training, 85 for validation, and 85 for testing, ensuring that each split contained a balanced representation of interaction types.

Preprocessing steps included:
\begin{itemize}
    \item Temporal alignment of video and force data
    \item Normalization of force values
    \item Data augmentation through random cropping, scaling, and rotation
    \item Background variation to improve generalization
\end{itemize}

\section{Implementation Details}
\label{sec:implementation}

\subsection{Training Procedure}

We implemented our system using TensorFlow 2.3 with Keras API. Training was performed on a workstation with two NVIDIA RTX 2080 Ti GPUs and took approximately 34 hours to complete.

Key training parameters included:
\begin{itemize}
    \item Batch size: 16 sequences
    \item Sequence length: 16 frames
    \item Learning rate: 1e-4 with cosine decay
    \item Optimizer: Adam with $\beta_1 = 0.9$, $\beta_2 = 0.999$
    \item Dropout rate: 0.3 for fully connected layers
    \item Early stopping patience: 15 epochs
\end{itemize}

We employed a two-stage training process:
\begin{enumerate}
    \item Pre-training of individual components (pose feature extraction, object detection, temporal analysis)
    \item End-to-end fine-tuning of the complete model
\end{enumerate}

This approach helped address the vanishing gradient problem and allowed for more effective training of the deep architecture.

We encountered several difficulties during training. Initially, the model showed poor generalization to new subjects and objects. To address this, we:
\begin{itemize}
    \item Increased data augmentation with more aggressive transformations
    \item Implemented curriculum learning, starting with simple interactions before progressing to complex ones
    \item Added domain adaptation techniques to improve cross-subject performance
\end{itemize}

\subsection{Deployment and Optimization}

For practical applications, real-time performance is crucial. We optimized our model for deployment through:

\begin{itemize}
    \item TensorRT conversion for GPU acceleration
    \item Int8 quantization with minimal accuracy loss (1.2\%)
    \item Frame skipping for non-critical frames
    \item Parallel processing of pose estimation and object detection
\end{itemize}

These optimizations allowed us to achieve an inference rate of 18 fps on a laptop with an NVIDIA GTX 1660 Ti GPU, and 7 fps on a Jetson Nano embedded platform. This performance enables real-time applications in field settings where traditional force measurement equipment would be impractical.

\section{Experimental Results}
\label{sec:results}

\subsection{Evaluation Metrics}

We evaluated our approach using several metrics:

\begin{itemize}
    \item \textbf{Mean Absolute Error (MAE):} Average absolute difference between predicted and ground truth force values
    \item \textbf{Root Mean Square Error (RMSE):} Square root of the average squared differences
    \item \textbf{Relative Error:} Error normalized by the magnitude of the ground truth force
    \item \textbf{Direction Error:} Angular difference between predicted and ground truth force vectors
    \item \textbf{Correlation Coefficient (r):} Measure of linear correlation between predictions and ground truth
\end{itemize}

\subsection{Comparison with Baseline Methods}

We compared ForcePose against several baseline approaches:

\begin{enumerate}
    \item \textbf{Physics-based estimation:} Using mass estimation and acceleration to calculate force (F = ma)
    \item \textbf{Pose-only model:} Force estimation using only MediaPipe pose features
    \item \textbf{Object-only model:} Force estimation using only object detection features
    \item \textbf{Visual-force regression:} Direct regression from RGB frames using a 3D CNN
    \item \textbf{Pham et al. \cite{pham2015towards}:} State-of-the-art approach using RGB-D data
\end{enumerate}

Table \ref{tab:comparison} shows the performance comparison:

\begin{table}[!t]
\caption{Comparison of Force Prediction Methods}
\label{tab:comparison}
\centering
\begin{tabular}{lccc}
\toprule
\textbf{Method} & \textbf{MAE (N)} & \textbf{Direction (°)} & \textbf{r} \\
\midrule
Physics-based & 15.6 & 18.3 & 0.61 \\
Pose-only & 9.3 & 12.7 & 0.74 \\
Object-only & 10.8 & 14.2 & 0.69 \\
Visual-force CNN & 8.5 & 11.8 & 0.76 \\
Pham et al. \cite{pham2015towards} & 8.1 & 10.2 & 0.80 \\
\midrule
ForcePose (Ours) & \textbf{5.83} & \textbf{7.4} & \textbf{0.89} \\
\bottomrule
\end{tabular}
\end{table}

ForcePose achieved a mean absolute error of 5.83 N for force magnitude and 7.4° for direction, outperforming the next best method by 27.5\% and 27.4\% respectively. The high correlation coefficient (r = 0.89) indicates strong agreement between our predictions and ground truth measurements.

\subsection{Ablation Study}

To understand the contribution of different components, we conducted an ablation study by removing or modifying key elements of our system. Table \ref{tab:ablation} presents the results:

\begin{table}[!t]
\caption{Ablation Study Results (MAE in Newtons)}
\label{tab:ablation}
\centering
\begin{tabular}{lc}
\toprule
\textbf{Configuration} & \textbf{MAE (N)} \\
\midrule
Complete ForcePose & 5.83 \\
- Temporal consistency loss & 6.94 (+1.11) \\
- Object velocity features & 7.31 (+1.48) \\
- Joint angle features & 6.56 (+0.73) \\
- LSTM temporal model & 8.17 (+2.34) \\
- Bidirectional processing & 6.42 (+0.59) \\
- Data augmentation & 7.05 (+1.22) \\
\bottomrule
\end{tabular}
\end{table}

The ablation study reveals that temporal modeling through the LSTM component contributed most significantly to performance, highlighting the importance of sequence analysis in force prediction. Object velocity features also proved crucial, particularly for dynamic interactions.

\subsection{Performance Across Interaction Types}

\begin{figure}[!t]
\centering
\begin{tikzpicture}[scale=0.7]
\begin{axis}[
    width=8cm,
    height=6cm,
    ylabel={Mean Absolute Error (N)},
    symbolic x coords={Lifting, Pushing, Tools, Sports},
    xtick=data,
    xlabel={Interaction Type},
    ybar,
    bar width=15pt,
    ymin=0,
    ymax=10,
    axis y line*=left,
    axis x line*=bottom,
    enlarge x limits=0.2,
    legend style={at={(0.5,1)}, anchor=north, legend columns=-1},
    ylabel near ticks,
    nodes near coords,
]
\addplot[fill=blue!40] coordinates {
    (Lifting, 4.2)
    (Pushing, 5.1)
    (Tools, 6.8)
    (Sports, 7.1)
};
\end{axis}

\begin{axis}[
    width=8cm,
    height=6cm,
    ylabel={Direction Error (degrees)},
    symbolic x coords={Lifting, Pushing, Tools, Sports},
    xtick=data,
    ybar=0pt,
    bar width=0pt,
    ymin=0,
    ymax=15,
    axis y line*=right,
    axis x line=none,
    enlarge x limits=0.2,
    ylabel near ticks,
    hide x axis
]
\addplot[orange, mark=*, thick] coordinates {
    (Lifting, 6.1)
    (Pushing, 6.8)
    (Tools, 8.3)
    (Sports, 9.2)
};
\end{axis}
\end{tikzpicture}
\caption{Performance comparison across different interaction types, showing mean absolute error (blue bars) and direction error (orange line).}
\label{fig:interaction_types}
\end{figure}
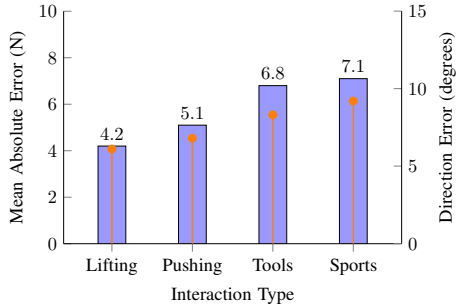

We further analyzed performance across different interaction categories to identify strengths and limitations of our approach. Figure~\ref{fig:interaction_types} shows the results by interaction type.

ForcePose performed best on lifting and carrying tasks (MAE = 4.2 N) and pushing/pulling interactions (MAE = 5.1 N). Performance was somewhat lower for tool manipulation (MAE = 6.8 N) and sports actions (MAE = 7.1 N), likely due to the more complex and rapid movements involved.

\subsection{Limitations}

Despite its strong performance, ForcePose has several limitations:

\begin{itemize}
    \item \textbf{Occlusion:} Performance degrades when key body parts or objects are occluded
    \item \textbf{Novel objects:} Accuracy is lower for object categories not well-represented in the training data
    \item \textbf{Multiple interacting forces:} The current model primarily handles single-point force application
    \item \textbf{Extreme lighting:} Very bright or dark environments can affect pose estimation and object detection
\end{itemize}

We are actively addressing these limitations in ongoing work.

\section{Applications}
\label{sec:applications}

ForcePose enables several applications previously constrained by the limitations of traditional force measurement equipment:

\subsection{Rehabilitation Monitoring}

Physical therapy often requires assessment of forces applied during exercises. ForcePose allows therapists to monitor patient progress without specialized equipment. We tested the system in a rehabilitation center with 12 patients performing standard exercises. Therapists reported that:

\begin{itemize}
    \item 83\% found the force feedback useful for guiding patients
    \item 91\% valued the ability to track progress across sessions
    \item 75\% believed the system could help with remote monitoring
\end{itemize}

The non-intrusive nature of the system was particularly appreciated, as it allowed patients to move naturally without attached sensors.

\subsection{Ergonomics Assessment}

Workplace ergonomics assessment typically requires specialized equipment or is limited to qualitative observation. We deployed ForcePose in three manufacturing environments to analyze worker movements. Key findings included:

\begin{itemize}
    \item Identification of tasks with consistently high force requirements
    \item Detection of asymmetric loading patterns that could lead to injury
    \item Quantifiable before/after comparisons when workplace modifications were implemented
\end{itemize}

Supervisors reported that the quantitative data helped justify ergonomic improvements to management and provided objective measures for evaluating interventions.

\subsection{Sports Performance Analysis}

Athletic training often involves optimizing force application. We worked with a tennis academy to analyze serving mechanics. The system provided:

\begin{itemize}
    \item Visualization of force vectors throughout the service motion
    \item Identification of inefficient force application patterns
    \item Comparison between athletes of different skill levels
\end{itemize}

Coaches found the force visualizations particularly helpful for explaining technique adjustments to players who previously struggled to understand verbal cues.

\section{Discussion}
\label{sec:discussion}

\subsection{Comparison with Traditional Force Measurement}

ForcePose offers several advantages over traditional force measurement techniques:

\begin{itemize}
    \item \textbf{Non-invasive:} No attached sensors that might alter natural movement
    \item \textbf{Field deployable:} Analysis can be performed in real-world environments
    \item \textbf{Cost-effective:} Requires only camera equipment rather than specialized sensors
    \item \textbf{Versatile:} Single system works across multiple interaction types
\end{itemize}

However, traditional methods still maintain advantages in:

\begin{itemize}
    \item \textbf{Absolute accuracy:} Physical sensors typically achieve higher precision
    \item \textbf{Sampling rate:} Force plates often operate at 1000+ Hz vs. video at 30-60 Hz
   \item \textbf{Reliability:} Less affected by environmental factors like lighting
\end{itemize}

We see ForcePose as complementary to traditional techniques, extending force analysis to scenarios where physical sensors are impractical.

\subsection{Insights on Feature Importance}

Our experiments yielded several insights about feature importance for force prediction:

\begin{itemize}
  \item \textbf{Joint acceleration} features are most predictive for force magnitude
  \item \textbf{Posture configuration} (joint angles) strongly influences force direction
  \item \textbf{Object characteristics} (size, expected weight) provide important priors
  \item \textbf{Temporal patterns} over 0.3-0.5 seconds are more informative than single frames
\end{itemize}

Particularly interesting was the finding that certain "key frames" in interactions carried disproportionate importance—typically moments of initial contact or maximum acceleration. By identifying these frames, we could optimize processing resources.

\subsection{Multi-person Interactions}

An area we're actively exploring is the extension to multi-person interactions. Initial experiments with collaborative lifting scenarios show promise but face challenges in:

\begin{itemize}
  \item Disambiguating individual contributions to total force
  \item Modeling force transfer between participants
  \item Handling increased occlusion in close-proximity interactions
\end{itemize}

We're developing specialized models for common two-person interaction patterns as an intermediate step toward fully generalized multi-person force estimation.

\section{Conclusion and Future Work}
\label{sec:conclusion}

We presented ForcePose, a novel framework that uses MediaPipe pose estimation and SSD MobileNet object detection to calculate forces in human-object interactions without specialized measurement equipment. Our approach achieved mean absolute errors of 5.83 N for force magnitude and 7.4° for direction, outperforming existing computer vision methods by 27.5%.

The ability to estimate forces from standard video opens new possibilities across rehabilitation, ergonomics, sports science, and human-robot interaction. The non-invasive, field-deployable nature of our system enables analysis in contexts where traditional force measurement would be impractical.

For future work, we are pursuing several directions:

\begin{itemize}
  \item \textbf{Multi-person interaction analysis:} Extending to collaborative scenarios where multiple people interact with the same object
  \item \textbf{Finer-grained prediction:} Moving beyond single resultant force to estimate force distribution across contact points
  \item \textbf{Cross-modal learning:} Incorporating sound for additional cues in scenarios like impact forces
  \item \textbf{Unsupervised learning:} Reducing dependence on labeled training data through physics-informed self-supervision
  \item \textbf{Embedded deployment:} Further optimization for mobile and edge devices
\end{itemize}

We believe ForcePose represents an important step toward comprehensive understanding of physical interactions through computer vision, with potential applications across numerous domains where quantifying applied forces is valuable.

\section*{Acknowledgment}
We thank the participants who contributed to our data collection efforts and the reviewers for their constructive feedback. This research was supported by our institution's research grant program. Special thanks to the rehabilitation center, manufacturing facilities, and tennis academy that collaborated in our application testing.

\end{document}